# Enhancing Retrieval-Augmented LMs with a Two-stage Consistency Learning Compressor

Chuankai Xu[1], Dongming Zhao[2], Bo Wang[1] and Hanwen Xing[3],

[1]College of Intelligence and Computing, Tianjin University, Tianjin, China
[2]Artificial Intelligence Laboratory, China Mobile Communication Group Tianjin Co., Ltd.
[3]University College London, London, WC1E 6BT, United Kingdom
Chuankai_Xu@tju.edu.cn waitman_840602@163.com  bo_wang@tju.edu.cn
13902100085@qq.com

**Abstract.** Despite the prevalence of retrieval-augmented language models (RALMs), the seamless integration of these models with retrieval mechanisms to enhance performance in document-based tasks remains challenging. While some post-retrieval processing Retrieval-Augmented Generation (RAG) methods have achieved success, most still lack the ability to distinguish pertinent from extraneous information, leading to potential inconsistencies and reduced precision in the generated output, which subsequently affects the truthfulness of the language model's responses. To address these limitations, this work proposes a novel two-stage consistency learning approach for retrieved information compression in retrieval-augmented language models to enhance performance. By incorporating consistency learning, the aim is to generate summaries that maintain coherence and alignment with the intended semantic representations of a teacher model while improving faithfulness to the original retrieved documents. The proposed method is empirically validated across multiple datasets, demonstrating notable enhancements in precision and efficiency for question-answering tasks. It outperforms existing baselines and showcases the synergistic effects of combining contrastive and consistency learning paradigms within the retrieval-augmented generation framework.

## 1    Introduction

Retrieval-augmented generation (RAG) has emerged as a promising paradigm for enhancing language model performance in document-based tasks. RAG comprises three stages: Retrieval, Augmentation, and Generation. In the Retrieval stage, the corpus is divided into chunks, and vector indices are created using an encoder model. The Augmentation stage identifies and retrieves relevant chunks by comparing vector representations of the query and the indexed chunks. Finally, in the Generation stage, the model generates a response conditioned on the information from the retrieved chunks.

Recent advancements in RAG focus on refining each stage, including advanced encoding and indexing for improved corpus representation, sophisticated retrieval mechanisms to identify pertinent chunks, and novel generation strategies for accurate and coherent responses. However, noise in retrieved documents challenges the seamless integration of retrieval mechanisms with language models, particularly in controlling the relevance and faithfulness of the generated outputs.

To address these limitations, we introduce the Contrastive and Consistency Learning Post-Retrieval Compressor (C2LPRCom) RAG framework. C2LPRCom aims to enhance robustness against noise in retrieved documents while maintaining consistency with the original source. By incorporating contrastive and consistency learning, we enhance the model's ability to identify and utilize relevant information from retrieved documents, improving the quality and faithfulness of the generated outputs.

This paper contributes to post-retrieval processing in RAG by:

a)  Proposing a fine-grained extractive compressor based on contrastive learning, introducing a three-way data construction method to enhance localization accuracy during extraction by leveraging granular positive, semi-positive, and negative sample pairs.
b)  Introducing a two-stage consistency learning approach for constructing a lightweight post-retrieval information compressor in RAG frameworks. This method optimizes generated outputs to be relevant and faithful to retrieved information, employing a teacher-student paradigm to ensure high semantic similarity and robustness against noise.
c)  Conducting extensive experiments on three benchmark datasets, covering question-answering and language modeling tasks. Results demonstrate significant improvements in precision and inference efficiency, outperforming state-of- the-art baselines and highlighting the potential of fine-grained contrastive and two-stage consistency learning techniques in enhancing RAG performance.

## 2    Methodology

### 2.1    Task Definition

Suppose the original input to the large model for a user is a sequence of N words: $\mathbf{X}= \{x_1, x_2, x_3, ..., x_N \}$. K relevant documents $D = \{D_1, ...D_k\}$ (k=1,2,...K) have been retrieved from the data pool based on the query. We aims to train such a model $M_\theta$, which output a summary sequence $S$ of the shortest possible length L that contains all the relevant information from the K documents. The training data for $M_\theta$ is created using the outputs of a contrastive selector and a distillation module. $M_\theta$ is then fine-tuned on this crafted dataset using a two-stage consistency learning way to optimize its performance in producing informative summaries.

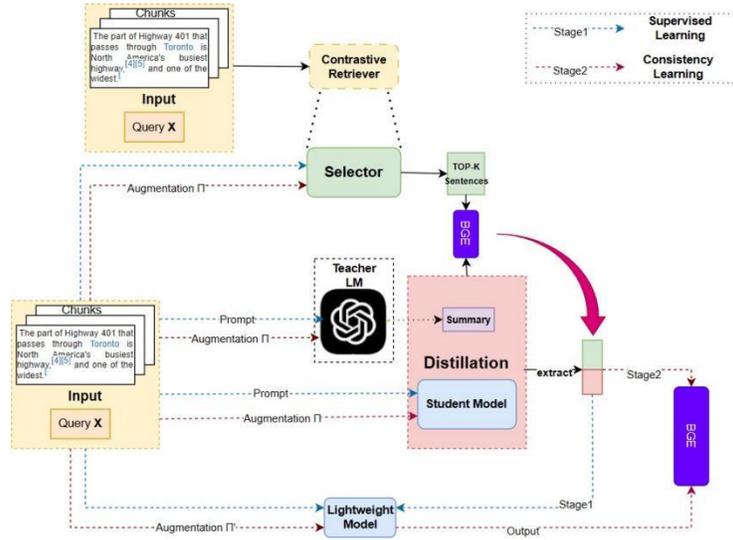

Fig. 1: Overview of our proposed C2LPRCom architecture

Our framework utilizes two complementary modules. The contrastive retriever enhanced selection model ranks and selects the optimal sentences $S_j$ from the retrieved document $D$ to form the sequence $S$. The distillation learning module takes the user's question and the retrieved relevant documents as input and outputs the best sequence $S_2$ as the summary. This summary is then incorporated with a prompt template to serve as the input for the subsequent large language model, which generates the final response to the original query $X$.

## 2.2 Contrastive Retriever on Sentence-Level

Some matured pre-trained dense retriever methods is prevailing and success- fully applied in some perplex language tasks, like Contriever[2], ColBert[3].But towards some specific tasks, they still need to be more fine-grained to escape introducing bias and noise, which could improve the performance of downstream task.

The training process of the encoder is similar to training common question- answering models. The loss used in model training takes the form of contrastive learning, similar to[4]:

$$L_c = -\sum \log \frac{\exp(\frac{sim(x, s_i)}{\tau})}{w_i \sum_{j=1}^{n} w_i \exp(\frac{sim(x, s_j)}{\tau})} \tag{1}$$

where $sim(x, s_i)$ is the inner product between embeddings, used to measure the similarity between sentences. $\tau$ is the temperature parameter used to normalize the similarity. $w_i$ is the weight assigned to each sample, with $w_i$ being 1 for positive and negative samples, and the value assigned in Algorithm 2 for semi-positive samples. The training objective of the model aims to minimize $L_c$ and maximize the distance between positive and negative samples, while semi-positive samples lie in between based on their weights. We have attached pseudo code in 2 in Appendix.

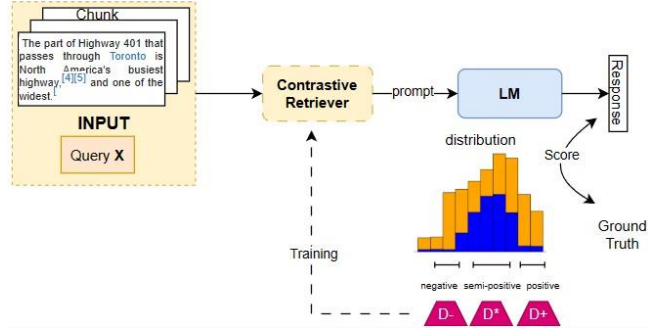

Fig. 2: Overview of our refined contrastive retriever

### 2.3  Distillation Module

The algorithm flow is shown in Algorithm 1, which illustrates the process of data distillation and subsequent consistency learning based on the outputs of the teacher and student models. In the first stage, we do not perform perturbation or data augmentation on the user input and retrieved results. For each user input $x_i$ and the retrieved documents $D_i$, we utilize a fine-tuned selector in the upstream to choose the top-k candidate sentences $CS(x_i, D_i)$. In the midstream distillation module, we obtain the corresponding output $S'$ for each template in $p_i$ by combining $x_i$ and $D_i$ as input to the teacher model $L'$. Simultaneously, we input whether to concatenate $S'$ with $x_i$ and $D_i$ to the student large language model $L$. If concatenating the teacher model output $S'$ helps the student model generate a reply with a higher score, we further utilize the teacher model output to construct the distilled training set; otherwise, we discard the teacher model output, indicating that the teacher large language model cannot perform better than the student model for that data point. We then calculate the similarity between the encoded $CS(x_i, D_i)$ and the teacher model output $S'$, and select the top-5 from the selector output as the result $s^e$, which is concatenated with the distiller output to form the constructed first-stage fine-tuning dataset $T_1$.

Similarly, in the second stage of fine-tuning data construction, we add different data augmentations to the inputs of the upstream selector, midstream distillation model, and downstream fine-tuning model. The remaining process is the same as in the first stage, except that we additionally require the fine-tuning model output under perturbation to maintain semantic consistency with the upstream and midstream constructed fine-tuning data $T_2$. During training, in the first M rounds, fully supervised fine-tuning is performed based on the first-stage constructed dataset. In subsequent rounds, in addition to the first-stage training loss, the consistency learning training loss is also added.

---

**Algorithm 1** Generate the Distilled Training Data

---

**Input:** lightweight model $M_\theta$, Student LM $L$, Teacher LM $L'$, Fine-tuned Selector $CS$, Input Training Data $\{X_i, D_i, y_i\}^T$, $X_i$ for User Input, Retrieved Relevant Document $D_i$, Ground Truth $y_i$, Data Augmentation Set $\Pi$, Prompt Template Set $\{pi\}^n$.

**Output:** Distillation Training Dataset $\{T_1, T_2\}$, $T_1$ for Stage 1, $T_2$ for Stage 2

$\qquad T_1, T_2 \rightarrow \varnothing, \pi, \pi' \in \Pi$

**for** $i = 1$ to T **do**

$\quad sc, sc' \rightarrow -\infty;\ s_t,\ s'_t,\ S_t^e,\ S_t^{e'}$

$\quad$ **for** $j = 1$ to n **do**

$\qquad ot_j, ot' \rightarrow L'(p_j; x_i; D_i),\ L'(p_j; \pi(x_i); \pi(D_i))$

$\qquad$ **if** $sc < SCORE(L, y_i, [ot_j; x_i])$ **then**

$\qquad\quad s_t \rightarrow ot_j; sc = SCORE(L, y_i, [ot_j; x_i])$

$\qquad$ **end if**

$\qquad$ **if** $sc' < SCORE(L, y_i, [ot'; \pi(x_i)])$ **then**

$\qquad\quad s'_t \rightarrow ot'_j; sc' = SCORE(L, y_i, [ot'; \pi(x_i)])$

$\qquad$ **end if**

$\quad$ **end for**

$\quad s_t^e \rightarrow argTop5s_j \in_{S_t^e} <Enc(CS(x_i, D_i)),\ Enc(L'(p_j; x_i; D_i)) >$

$\quad s_t^{e'} \rightarrow argTop5s_j \in_{S_t^e} <Enc(CS(\pi(x_i)),\ \pi(D_i))),\ Enc(L'(p_j; \pi(x_i)); \pi(D_i)))>$

$\quad$ **if** $sc < SCORE(L, y_i, [x_i])$ **then**

$\qquad T_1 \cup \{(x_i, D_i, s_t^e)\},\ T_2 \cup \{(\pi(x_i), \pi(D_i), s_t^{e'})\}$

$\qquad$ continue

$\quad$ **end if**

$\quad T_1 \cup \{(x_i, D_i, [s_t^e; s_i])\},\ T_2 \cup \{(\pi'(x_i), \pi'(D_i), [s_t^{e'}; s'_t])\}$

**end for**

**return** $\{T_1, T_2\}$

---

For the language modeling task, $SCORE(L, y_i, [s_j; x_i]) = log p_L(y_i/s_j; x_i)$, which represents the log-likelihood score of the output of the LLM $L$ with respect to the reference answer, given the concatenation of the selected best candidate sentence and the user input. In the case of question answering (QA) tasks, SCORE employs the Exact Match between strings as the evaluation metric.

### 2.4    Two-stage Consistency Learning

After obtaining the distilled training datasets $T_1$, $T_2$, a two-stage consistency learning approach is employed for model training. During the initial stage, the model undergoes fine-tuning solely guided by the supervision loss. Subsequently, in the second stage, the model is trained under the joint supervision of both the supervised loss and the consistency loss, enabling the model to learn from both labeled data and the consistency between the predictions of the model on different views of the input data.

The fully supervised loss $L_s$ employs cross-entropy loss to measure the discrepancy between the model-generated sequence and the ground-truth answer.

For a single sample sequence, the cross-entropy loss is defined as follows on a per word basis:

$$L_s(x_i, y_i) = -\sum_{j=1}^{n} logP(y_{ij} \mid x_i, y_{i1}, y_{i2}, ...., y_{(i(j-1))}) \tag{2}$$

$$L_s = 1/N \sum_{j=1}^{N} L(x_i, y_i) \tag{3}$$

For the consistency learning loss, it primarily consists of two components: By inputting the constructed training set $[s^{e'}; s']$ under perturbation and the output $M_\theta(\pi'(x_i, D_i))$ of the lightweight model $M_\theta$ into the BGE-en-large[8] encoder, we obtain their embedded representations. We then separately calculate the L2 norm between $s^{e'}$ and $M_\theta(\pi'(x_i, D_i))$, as well as between $s'$ and $M_\theta(\pi'(x_i, D_i))$, as the final loss. The formulation is as follows:

$$L_{cg} = 1/N \sum_{j=1}^{N} \| enc(s') - enc(M_\theta(\pi'(s_i, D_i))) \|_2^2 \tag{4}$$

$$L_{ce} = 1/N \sum_{j=1}^{N} \| enc(s_t^{e'}) - enc(M_\theta(\pi'(s_i, D_i))) \|_2^2 \tag{5}$$

The total training loss is expressed as, where $\lambda_1$, $\lambda_2$ are ramp-up weighting factor, controlling the trade-off between supervised loss and consistency loss. On first stage, they are set to zero :

$$L = L_s + \lambda_1 L_{cg} + \lambda_2 L_{ce} \tag{6}$$

## 3    Experimentation

### 3.1    Datasets and Measure Metrics

We conduct our experiment on three selected datasets for two tasks–Question Answering (Non-reasoning & Reasoning) and language modeling.

**WikiText-103[4]:** is extracted from Wikipedia articles, is designed for language model training and evaluation, focusing on long-term dependencies. It contains over 100 million tokens, with 101,425,671 tokens in the training set, 213,886 tokens in the validation set, and 241,211 tokens in the test set.

**Natural Question[5]:** a large-scale dataset for evaluating the natural language generation ability of models. It consists of a training set with 307,373 annotated question-answer pairs and a validation set and test set, each containing 7,830 and 7,840 5-way annotated data pairs, respectively.

**HotpotQA[6]:** aims to require models to perform question answering by integrating multi-hop reasoning and information from multiple documents. It includes a large number of questions, relevant documents, and correct answers, with approximately 90,000 training examples and 7,400 validation examples.

Following the metrics in work[7], we choose model output's **perplexity(PPL)** to evaluate the performance of model's language modeling ability, and **Exact Match(EM)** and **Token-level F1** to assess model's performance on QA task.

## 3.2 Hyperparameters and Other Settings

This study utilizes the NLTK package for sentence segmentation and filters out data pairs lacking negative or semi-positive examples. Document retrieval is performed using BM25[13]. For the distillation module, we employ the GPT-3.5-turbo[9] API with a temperature of 0.7 and Top p of 1. In the language modeling task, three prompts are used (as shown in the WikiText-103 data column), with the lowest perplexity result selected as the target answer. Approximately 2% of the training data (36,000 examples) are randomly sampled to generate three summaries per example.

The contrastive selector initializes $en_\theta$ with Contriever[4], fine-tuned on the MS Marco[10] dataset. With 110M parameters, it is optimized using Adam with a learning rate of 1e-5, 2000 warmup steps, and trained over 5 epochs with a batch size of 64. Evaluation results are selected based on validation set performance. In the language modeling task, GPT2 and GPT2-XL[9] are employed as the lightweight models, with GPT2-XL transferred directly from GPT2. For the QA task, LLaMA2 (13B)[11] serves as the student model, while GPT-3.5-turbo is the teacher model. The lightweight model uses T5-large[12] with 770M parameters for fine-tuning, following a similar Adam optimization strategy over 6 epochs. The first three epochs are fully supervised, followed by three epochs incorporating consistency learning, with a batch size of 32. BGE-en-large is used as the sentence encoder Enc.

For data augmentation, two main strategies are adopted: token-level augmentation (synonym replacement and random token insertion/deletion) and sentence-level paraphrasing[14]. Due to computational constraints, 30% of sentences from retrieved documents are randomly sampled for augmentation. Different augmentation types are employed for downstream fine-tuning and up- stream/midstream modules.

### 3.3    Baselines

For the language modeling task, there are two heuristic baseline models at the token and phrase levels:

- **Bag-of-Words (BoW):**Converting the retrieved documents into an ordered list of words and concatenates them;
- **Named Entity (NE):**Extracting an ordered list of named entities from the retrieved documents and concatenates them.
- **BM25:** We select it as one representation of classic retrieval algorithms.
- **Contriever:** Our initialization dense retrieve model, fine-tuned on MS MARCO.
- **Random:** We random select a sentence from retrieved documents.
- **RECOMP[7](extractive):**The SOTA post-retrieve RAG model by extractively summarizing the retrieved documents.
- **Upperbound:** Upperbound performance is obtained by traversing all the sentences in the retrieved document w.r.t input query X to achieve the best evaluation score.

For QA task, we adopt the following baselines:

- **GPT-turbo-3.5:**We directly get the response by inputting query X.
- **T5:**We use the off-the-shelf T5-large version.
- **RECOMP(abstractive):**The SOTA post-retrieve RAG model by abstractively summarizing the retrieved documents.

For the dynamic sampling selector in the language modeling task, we use BM25 and Contriever as baseline models, which rank sentences based on their similarity to the input x. For the question answering task, we report using BM25 and Contriever fine-tuned on the MS MARCO dataset as comparative baseline models.

### 3.4    Results and Analysis

In the language modeling task, Table 1 presents the performance of baseline models compared to our model. First, GPT2 and GPT2-XL perform worse without retrieval than with any retrieval-based input. Second, concatenation based on named entities (NE) performs worse than the bag-of-words (BoW) model, and both are inferior to directly inputting retrieved documents. This may be because BoW and NE provide less effective information.

Notably, when the selector module is active (using only the selector's retrieved results as a summary), there is a significant difference in average token length between the upper bound scheme and generated methods. Our method shows a smaller gap in average token length between the two modules, indicating better compression of original retrieved documents and more refined output, improving training and inference efficiency.

Compared to SOTA models like GPT-3.5 and RECOMP, our proposed modules show improved output length and model perplexity. For instance, the consistency learning module outperforms using the entire document for retrieval-augmented generation, indicating protection from noise in retrieved documents. In the question-answering task, as shown in Table 2, all retrieval-augmented methods improve EM and F1 metrics compared to non-retrieval methods, consistent with previous studies. Unlike the language modeling task, appending five documents as retrieval enhancement yields more significant improvements than a single document, indicating that more effective context enhances model performance. Across all datasets, the upper bound performance of the selector model surpasses the generation model. The upper bound performance is calculated by choosing the best from N candidate sentences for the selector model, while the generation model selects the optimal result from the teacher and student large language model outputs.

Table 1: Performance of automatic sampling contrastive model

| | GPT2(117M) | | GPT2-XL(1.5B) | |
|---|---|---|---|---|
| | tokens | PPL | tokens | PPL |
| Zero-shot | 0 | 37.84 | 0 | 19.89 |
| Upperbound | 32 | 30.61 | 32 | 16.69 |
| Prepend Document(TOP1) | 147 | 32.99 | 147 | 18.11 |
| Prepend Document(TOP5) | 521 | 35.89 | 521 | 19.65 |
| BoW | 67 | 36.27 | 66 | 18.99 |
| NE | 33 | 37.37 | 35 | 19.75 |
| **Extractive models** | | | | |
| BM25 | 33 | 36.87 | 33 | 19.23 |
| Contriever | 34 | 35.78 | 34 | 19.09 |
| Random | 27 | 37.01 | 27 | 19.63 |
| RECOMP(extractive) | 31 | 33.67 | 31 | 18.19 |
| **Ours(w/o distill module)** | **30** | **33.12** | **30** | **18.01** |
| **Generated models** | | | | |
| GPT-3.5 | 35 | 34.98 | 35 | 18.82 |
| T5 | 16 | 37.92 | 16 | 19.98 |
| RECOMP(abstractive) | 16 | 33.68 | 16 | 18.19 |
| **Ours(w/o selector)** | **15** | **33.11** | **30** | **17.99** |
| **Ours** | **17** | **32.98** | **31** | **17.87** |

The trained summary retrieval-augmented generation modules, both the selector and distillation modules, show performance improvements. On the NQ dataset, the selector module achieves a 5% compression rate while losing 2 EM points compared to appending the complete document. On the HotpotQA dataset, which requires multi-step document understanding, the selector module achieves an 11% compression rate while losing 2.4 EM points. Further research is needed for summarization in complex tasks like HotpotQA, as large language models, though competitive in single-document summarization, struggle with synthesizing information from multiple documents.

We use the number of tokens in the generated summary and inference speed, measured by GPU time, as indicators of model efficiency. Table 3 reports the GPU time for LLaMA2 (13B) on the NQ dataset, run on six RTX 4090 GPUs. For generating retrieval-augmented summaries, RECOMP and our method were run on two RTX 4090 GPUs, with the consistency module using T5-large and the dynamic sampling module using Contriver. Our method significantly improves efficiency compared to appending the top-5 documents, even when accounting for summary generation time. The dynamic sampling module is particularly efficient. Inference speed is influenced more by model parameter size and the benchmark model than by token count. For example, the consistency learning module (770M parameters) incurs more latency than the dynamic sampling module (110M parameters). Both of our modules demonstrate significant improvements in inference efficiency compared to RECOMP, with a 9% to 16% improvement rate.

Table 2: Performance of Seletor and Distillation Module on QA task.

| | NQ avg # token | F1 | EM | HotpotQA avg # token | F1 | EM |
|---|---|---|---|---|---|---|
| Zero-shot | 0 | 29.67 | 22.05 | 0 | 26.43 | 17.95 |
| Prepend Document(TOP1) | 135 | 31.23 | 23.72 | 141 | 40.66 | 28.87 |
| Prepend Document(TOP5) | 465 | 36.28 | 28.23 | 689 | 43.34 | 32.31 |
| BoW | 459 | 36.12 | 27.89 | 264 | 35.76 | 25.21 |
| NE | 344 | 30.88 | 23.13 | 165 | 31.54 | 22.12 |
| **Extractive models** | | | | | | |
| Random | **33** | 30.78 | 23.10 | **62** | 29.52 | 20.87 |
| BM25 | 37 | 33.43 | 25.45 | 75 | 37.82 | 26.72 |
| Contriever | 37 | 31.54 | 29.81 | 79 | 39.12 | 28.12 |
| RECOMP(extractive) | 38 | 43.92 | 36.41 | 76 | 39.23 | 28.13 |
| **Ours(w/o distill module)** | 36 | **44.74** | **37.12** | 72 | 39.91 | 28.83 |
| **Generated models** | | | | | | |
| GPT-turbo-3.5 | 57 | 46.14 | 36.89 | 110 | 39.89 | 30.12 |
| T5 | **11** | 34.42 | 25.71 | **8** | 33.01 | 23.01 |
| RECOMP(abstractive) | 37 | 45.16 | 36.92 | 66 | 37.51 | 28.04 |
| **Ours(w/o selector)** | 30 | **46.33** | **37.76** | 59 | **38.02** | 28.79 |
| **Ours** | 31 | **46.53** | **37.93** | 62 | **38.39** | **28.95** |

Table 3: The inference efficiency under different circumstances

| Model | token | Generated Time | Inference Time | Total Time |
|---|---|---|---|---|
| No context | 0 | 0 | 3988s | 3988s |
| Append TOP5 retrieved docs | 465 | 0 | 9891s | 9891s |
| RECOMP | 37 | 898s | 4561s | 5459s |
| Ours(w/o distill) | 30 | 783s | 4319s | 5002s |
| Ours(w/o selector) | 30 | 101s | 4117s | 4218s |

# 4    Conclusion

In conclusion, this paper proposes a novel two-stage consistency learning frame- work, C2LPRCom, to enhance the robustness of retrieval-augmented language models (RALMs) against noise in retrieved documents while maintaining consistency with the original source information. The framework incorporates a fine-grained extractive compressor based on contrastive learning and a two-stage consistency learning approach for constructing a lightweight post-retrieval information compressor. Extensive experiments on question-answering and language modeling tasks demonstrate that the proposed method significantly improves the precision and inference efficiency of RAG frameworks, outperforming state-of-the-art baselines. The synergistic combination of contrastive and consistency learning paradigms within the retrieval-augmented generation framework shows great potential for enhancing the performance of RALMs. Future research could explore further improvements to summary generation models for complex tasks and investigate the synthesis of information from multiple documents.

**Acknowledgements.** This work was supported by the National Key R&D Program of China(2022YFC3301900;2022YFC3301901), National Natural Science Foundation of China(62376188,62272340,62276187,62376192).

# 1.A    Appendix

---

**Algorithm 2** Contrastive Retriever Training Process

---

**Input:** encoder $en_\theta$, LM $L$, Input Data$\{x_i, S_i, y_i\}^T$, User Input $X_i, S_i = \{s^j\}^n$ Ground Truth $y_i$, Threshold $K, \delta$.

**Output:** contrastive fine-tuned retriever $en_\theta$

    $Ne, T \to \varnothing$

    **for** $i = 1$ to T **do**

        $sc \to \varnothing$

        $top_i \to argmax_{s_j \in \{S_i\}}SCORE(L, y_i, [s_j; x_i])$

        **for** $j = 1$ to n **do**

            $p_i \to < en_\theta(x_i), en_\theta(s_j) >; sc \to sc \cup \{p_i, s_j\}$

        **end for**

        $sc \to DESCENDINGSort(sc)[: K]$according to$p_i$

        **for** $m = 1$ to K **do**

            $\Delta p_i = p_m - p_1$

            **if** $\Delta p_i < \delta$ **then**

                $w_m \to 1/\Delta p_i$

                $Ne \to Ne \cup \{x_i, top_i, s_m w_m\}$

            **else**

                $T \to T \cup \{x_i, top_i, s_m\}$

            **end if end for**

**end for**

$en_\theta = $ Finetune(T, $en_\theta$)

**return** $en_\theta$

---